\documentclass[lettersize,journal]{IEEEtran}
\usepackage{amsmath,amsfonts}
\usepackage{algorithmic}
\usepackage{algorithm}
\usepackage{array}
\usepackage[caption=false,font=normalsize,labelfont=sf,textfont=sf]{subfig}
\usepackage{textcomp}
\usepackage{stfloats}
\usepackage{url}
\usepackage{verbatim}
\usepackage{graphicx}
\usepackage{cite}
\usepackage{amssymb}
\begin{document}

\title{Constructed regional photovoltaic generation fields improve ultra-short-term distributed solar forecasting}
\author{
Dayin~Chen,
Xiaodan~Shi,
Xin~Xu,
Zhongyue~Su,
Shibo~Zhu,
Dongxiao~Zhang,
Yuntian~Chen,
Jinyue~Yan%
\thanks{
D. Chen, Z. Su, D. Zhang, and Y. Chen are with the Zhejiang Key Laboratory of Industrial Intelligence and Digital Twin, Eastern Institute of Technology, Ningbo 315200, China.
}
\thanks{
D. Chen, Z. Su, D. Zhang, and Y. Chen are also with the Ningbo Key Laboratory of Advanced Manufacturing Simulation, Eastern Institute of Technology, Ningbo 315200, China (e-mail: dzhang@eitech.edu.cn; ychen@eitech.edu.cn).
}
\thanks{
D. Chen, S. Zhu, X. Xu, and J. Yan are with the Department of Building Environment and Energy Engineering, and the International Centre of Urban Energy Nexus, The Hong Kong Polytechnic University, Hong Kong SAR, China.
}
\thanks{
X. Shi is with the Department of Computer and Systems Sciences, Stockholm University, Stockholm, Sweden (e-mail: xiaodan.shi@su.se).
}
\thanks{
Corresponding authors: Xiaodan Shi, Yuntian Chen, and Jinyue Yan (e-mail: xiaodan.shi@su.se; ychen@eitech.edu.cn; j-jerry.yan@polyu.edu.hk).
}
}


\maketitle

\begin{abstract}
Ultra-short-term forecasting of densely distributed photovoltaic (PV) systems remains challenging because external meteorological and satellite products may not adequately resolve rapid local variations, while heterogeneous PV systems exhibit different generation responses under similar atmospheric conditions. Meanwhile, dense PV fleets provide spatially distributed measurements that reveal regional generation dynamics. Here, we reformulate fleet-level PV forecasting from modeling discrete station time series to predicting explicit, spatially structured regional PV generation fields. We develop SatPV-Field, a regional-to-local forecasting framework that transforms distributed PV measurements into image-like generation fields and models their spatiotemporal evolution. By combining historical PV fields with satellite-derived future shortwave radiation guidance and meteorological forecasts, the framework predicts future regional generation states, which subsequently provide location-specific guidance for individual-station forecasting. Experiments involving 496 distributed PV systems, including 446 source stations and 50 held-out target stations, demonstrate that these regional fields can be effectively forecast and used to improve station-level predictions. The proposed approach reduces station-level MAE by approximately 10.0\% relative to the best-performing single-station temporal baseline and by 2.5\% relative to a graph-based regional forecasting approach using the same source fleet. These findings establish an explicit spatial-field modeling paradigm that connects regional generation dynamics with individual-site forecasting.
\end{abstract}


\section{Introduction}

Solar photovoltaic (PV) deployment is expanding rapidly and becoming increasingly distributed, with global cumulative capacity approaching 3 TW in 2025 and distributed PV capacity in China reaching 530 GW \cite{R1,R2}. The proliferation of rooftop, commercial, and industrial PV systems is shifting the forecasting problem from individual plants toward fleets of geographically dispersed generators \cite{new2}. Unlike an isolated PV plant, a distributed fleet experiences spatially varying atmospheric conditions and exhibits generation dynamics that evolve across both space and time \cite{new5}. Reliable ultra-short-term forecasting of these dynamics is increasingly important for distribution-level visibility and local grid operation \cite{GEIT_chen,paper5,new7}. Distributed PV forecasting is therefore becoming not only a temporal prediction problem at individual sites, but also a problem of characterizing generation evolution across a region.

A fundamental challenge in distributed PV forecasting is the limited availability of weather observations at the spatial and temporal scales of individual sites. Large utility-scale PV plants can be equipped with dedicated meteorological sensors or sky-imaging systems to obtain high-resolution local information, whereas extending such sensing infrastructure to large numbers of small distributed PV systems is costly and often impractical \cite{paper5,erdener2022review}. Distributed PV forecasting therefore commonly relies on externally available numerical weather prediction (NWP) and satellite-derived products \cite{chen2025selecting,multimodal_review4}. Although these products provide valuable regional atmospheric information, their ability to resolve local conditions at individual PV installations remains limited. Satellite observations provide spatially detailed information on recent cloud and irradiance variations, but do not directly describe future atmospheric conditions. In contrast, NWP forecasts provide information on future weather evolution, yet may struggle to capture rapid, localized fluctuations relevant to ultra-short-term PV forecasting \cite{R7,new10}. Moreover, the spatial and temporal resolutions of both products may not fully resolve the fine-scale variability associated with densely distributed PV systems \cite{new11}. Beyond these limitations, even an accurate description of local atmospheric and irradiance conditions does not uniquely determine PV generation. Systems exposed to similar solar resources can exhibit different outputs because their irradiance-to-power responses depend on site-specific characteristics, including system configuration, module orientation and tilt, shading, conversion efficiency, degradation, and operating conditions \cite{R8,new_ref3,R9,R9_2}. Consequently, external meteorological information alone cannot fully characterize the heterogeneous spatial patterns of actual PV generation.

\begin{figure*}[t]
\centering
\includegraphics[width=2\columnwidth]{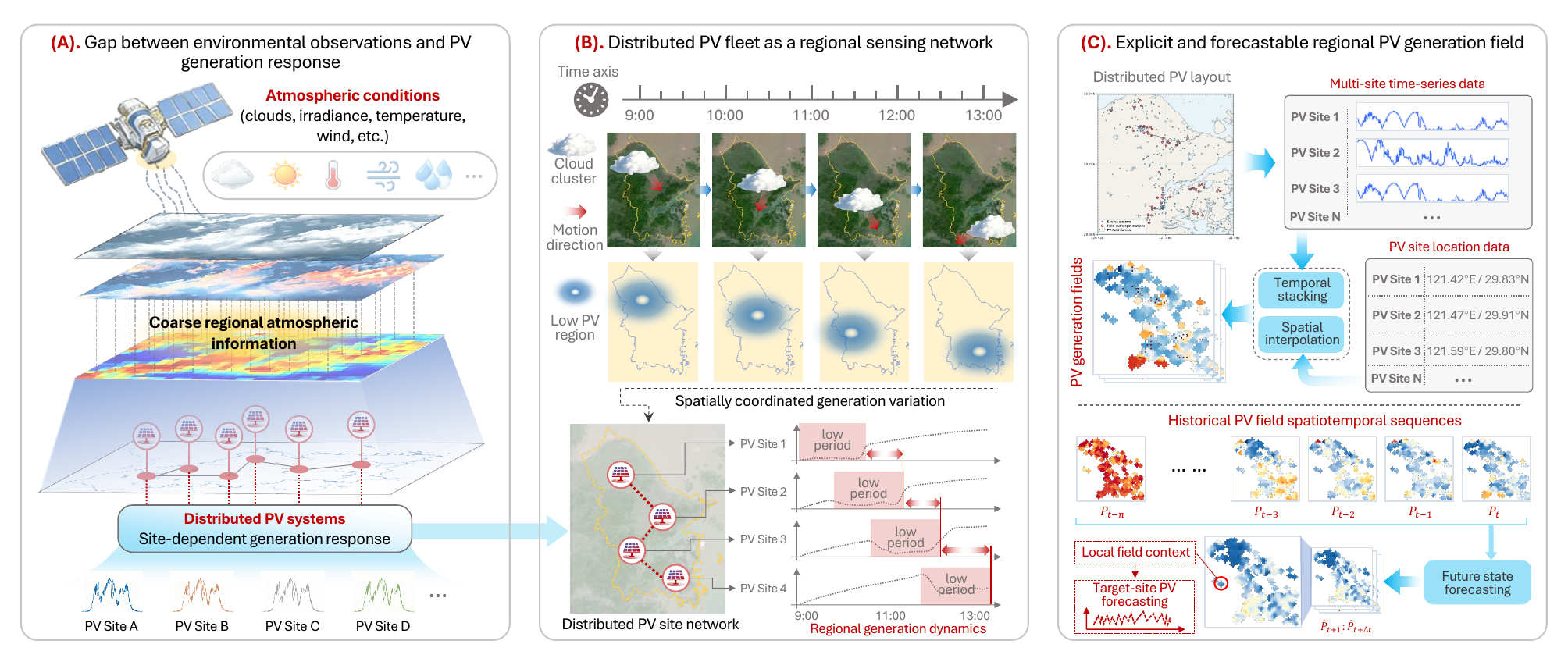}
\caption{
Conceptual motivation for representing distributed PV fleets as an explicit regional generation state.
(A) Atmospheric observations describe external conditions but cannot fully determine heterogeneous PV generation responses.
(B) Distributed PV systems provide spatially distributed measurements that form a regional sensing network capturing generation dynamics.
(C) Multi-site PV measurements can be transformed into an explicit and forecastable regional PV generation field, which provides future regional context for downstream station-level forecasting.
}
\label{fig:conceptual_motivation}
\end{figure*}

The same dense deployment that complicates site-level forecasting also creates a distributed sensing resource through PV generation measurements collected across the fleet \cite{sup1, sup4}. Unlike dedicated meteorological observations, which are often unavailable at individual small-scale PV installations, power measurements are routinely collected during system operation. These measurements reflect local generation responses to prevailing atmospheric conditions, shaped by the characteristics of individual PV systems. When combined across geographically distributed sites, they provide spatially distributed observations of the evolving regional generation state \cite{new8}. As weather-driven changes in cloud cover and irradiance affect different sites with spatial and temporal offsets, the collective measurements reveal regional generation dynamics that cannot be captured by any single site's history \cite{sup2}. Dense PV fleets can therefore be viewed not merely as collections of forecasting targets, but as regional sensing networks for observing the spatiotemporal evolution of PV generation. Fig.~\ref{fig:conceptual_motivation} illustrates this perspective, highlighting how distributed PV measurements can reveal regional generation patterns beyond those characterized by external atmospheric information alone.

Recent studies have explored the use of measurements from multiple PV sites to improve forecasting at individual stations. Neighboring PV systems provide information about correlated generation fluctuations and their temporal offsets across locations, helping characterize regional variations that may not be evident from a single site's history \cite{R11_1,R11_2,multisite1,multisite2}. Multi-site and graph-based approaches incorporate these spatial dependencies through geographical coordinates, graph structures, or learned spatial features \cite{new3,sup3}. In most of these approaches, cross-site information is used to support predictions at individual stations, while the regional generation state itself is not explicitly represented or forecast.

A prior study provided an initial exploration of spatializing distributed PV measurements by constructing two-dimensional solar PV maps from geographically distributed power data \cite{PV_field_first}. The study mainly focused on demonstrating the feasibility of this spatial representation and reported only a preliminary forecasting experiment, which the authors themselves described as very early work. The broader potential of such spatialized PV representations as a forecastable regional generation state for downstream site-level prediction was not systematically investigated.

These developments motivate a different formulation of fleet-level PV forecasting. Rather than modeling distributed PV systems solely as discrete station time series, we organize measurements from the source fleet into spatially structured regional PV generation fields. Stacking these fields over time transforms the station-indexed observations into a spatiotemporal field sequence, allowing geographical relationships, local generation patterns, and their temporal evolution to be modeled jointly. This representation makes the regional generation state explicit and directly accessible to spatial and spatiotemporal forecasting models. Unlike atmospheric or irradiance fields, the PV generation field captures the realized generation responses of distributed PV systems under varying weather conditions and site-specific characteristics, providing a representation more directly aligned with PV forecasting tasks. We therefore investigate whether predicting the evolution of this regional state can provide complementary spatial information beyond a target site's own history and meteorological inputs, thereby improving station-level predictions.

\begin{figure*}[t]
\centering
\includegraphics[width=\textwidth]{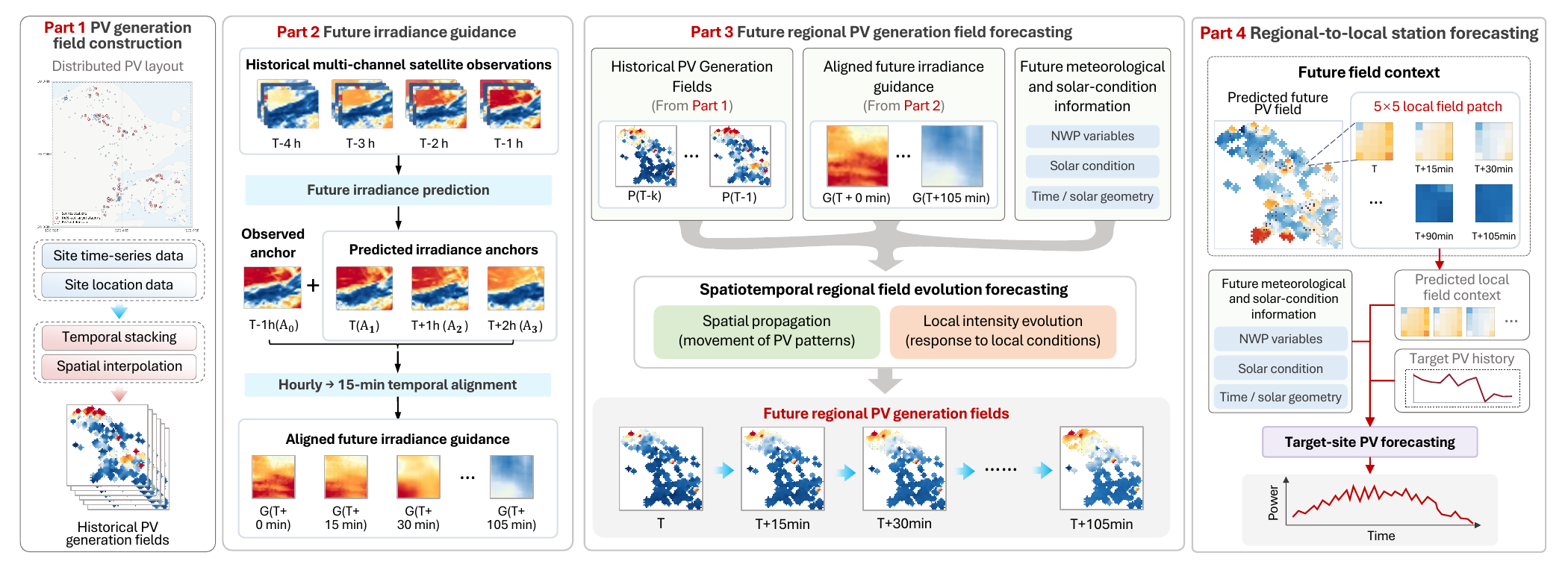}
\caption{
Overview of SatPV-Field.
The framework consists of four stages:
(1) construction of historical regional PV generation fields from distributed PV measurements;
(2) generation of future irradiance guidance from satellite observations;
(3) forecasting of future regional PV generation fields by integrating historical field dynamics, future irradiance guidance, and meteorological information;
and (4) regional-to-local station forecasting using predicted local field context together with target-site historical PV and future meteorological information.
}
\label{fig:framework}
\end{figure*}

To test this hypothesis, we develop SatPV-Field, a regional-to-local forecasting framework that predicts future regional PV generation fields and uses them to guide individual-station forecasting, as illustrated in Fig.~\ref{fig:framework}. First, historical satellite observations are used to predict future shortwave radiation fields, providing guidance on upcoming atmospheric conditions. These irradiance forecasts are then combined with historical regional PV generation fields and future meteorological forecasts to predict the spatiotemporal evolution of regional PV generation. Finally, location-specific context is extracted from the predicted PV fields and combined with each target site's historical PV observations and future meteorological information to forecast its power output.

To isolate the contribution of regional fleet information, the evaluation explicitly separates target-local information from the regional representation. Each target site retains its own historical PV observations and meteorological inputs for local forecasting, while its PV measurements are excluded from regional-field construction and field-level supervision. The regional generation field is therefore derived solely from the surrounding source fleet. This design avoids recycling target-site measurements through the regional representation and allows us to assess whether a forecasted regional generation state provides predictive information beyond that already available from the target site itself.

The main contributions of this work are summarized as follows:
\begin{itemize}

\item We formulate distributed PV generation as an explicit regional field forecasting problem. By transforming discrete measurements from a source fleet into spatially structured generation fields, the proposed formulation enables geographical relationships and the spatiotemporal evolution of regional PV generation to be modeled directly.

\item We develop SatPV-Field, a regional-to-local forecasting framework that combines satellite-derived future irradiance forecasts, historical PV generation fields, and meteorological forecasts to predict future regional generation states. The predicted fields provide location-specific spatial context for downstream station-level forecasting.

\item We evaluate the proposed framework using 496 real-world distributed PV systems, with 446 source stations and 50 held-out target stations. The experiments demonstrate the effectiveness of predicted regional field guidance, achieving approximately 10.0\% lower station-level MAE than the best-performing single-station temporal baseline and 2.5\% lower MAE than a matched graph-based regional forecasting approach, with additional evaluations across source densities and operating conditions.

\end{itemize}

\section{Methodology}
\label{sec:method}

\subsection{Problem Formulation: Explicit Regional PV Generation Field Forecasting}
\label{sec:problem_formulation}

We consider a set of geographically distributed source PV stations $\mathcal{S}$ and a disjoint set of target stations $\mathcal{T}$. Rather than modeling the source fleet solely as discrete station time series, we organize its power measurements into an explicit regional PV generation field. At each time step $t$, the field is constructed as
\begin{equation}
\mathbf{P}_{t}
=
\Phi\left(
\{p_i(t),x_i,y_i\}_{i\in\mathcal{S}}
\right),
\end{equation}
where $\mathbf{P}_t\in\mathbb{R}^{H_g\times W_g}$ denotes the spatially structured PV generation field, with $H_g=W_g=64$ in this study, and $(x_i,y_i)$ denotes the geographical coordinates of source station $i$. A corresponding support mask $\mathbf{M}_t$ identifies grid cells supported by source-station observations.

The first forecasting objective is to predict the future evolution of the regional generation field:
\begin{equation}
\begin{aligned}
\widehat{\mathbf{P}}_{T:T+H-1}
= f_{\mathrm{field}}\big(
&\mathbf{P}_{T-k:T-1},
\mathbf{M}_{T-k:T-1},\\
&\widehat{\mathbf{R}}_{T:T+H-1},
\mathbf{Z}_{T:T+H-1}
\big).
\end{aligned}
\end{equation}
where $\mathbf{P}_{T-k:T-1}$ and $\mathbf{M}_{T-k:T-1}$ denote historical regional PV fields and their support masks, $\widehat{\mathbf{R}}_{T:T+H-1}$ represents predicted future irradiance guidance, and $\mathbf{Z}_{T:T+H-1}$ represents future meteorological and solar-condition information. Here, \(k\) and \(H\) denote the numbers of historical and future time steps, respectively. For compactness, sequence subscripts use discrete 15-min time steps, with \(T-1\) corresponding to \(T-\Delta t\), where \(\Delta t=15\) min.

The predicted regional fields are subsequently used to guide forecasting at individual target stations:
\begin{equation}
\begin{aligned}
\widehat{\mathbf{p}}^{\,\mathrm{pu}}_{j,T:T+H-1}
=
f_{\mathrm{station}}\big(
&\mathbf{X}^{\mathrm{hist}}_{j,T},
\mathbf{Z}^{\mathrm{fut}}_{j,T},\\
&\mathcal{C}_j(
\widehat{\mathbf{P}}_{T:T+H-1},
\mathbf{M}_{T-\Delta t}
)
\big).
\end{aligned}
\end{equation}
where $\mathcal{C}_j(\cdot)$ extracts the location-specific field context and support mask around target station $j$. The station forecaster combines this regional context with the target's own historical observations and future meteorological information to predict its power output.

\subsection{Source-Supported Regional PV Generation Field Construction}
\label{sec:field_construction}

The regional PV generation field is constructed exclusively from source-station measurements, with target stations excluded from field construction and supervision.

For source station $i\in\mathcal{S}$ with installed capacity $C_i$ and measured power output $p_i(t)$, the power measurement is first capacity-normalized and then rescaled using a station-specific upper quantile:
\begin{equation}
\tilde{p}_i(t)
=
\operatorname{clip}
\left(
\frac{p_i(t)/C_i}{q_{i,0.95}},
0,p_{\max}
\right),
\label{eq:pv_normalization}
\end{equation}
where $q_{i,0.95}$ is the 0.95 quantile of the capacity-normalized power at station $i$, estimated using the training split only, and $p_{\max}=1.5$ is used to suppress extreme values. Capacity normalization removes differences in installed system size, while station-specific quantile scaling further aligns generation magnitudes across the source fleet without removing their temporal variations.

The study region is discretized into a $64\times64$ regular grid. After mapping each source station to its geographical grid location, the normalized measurement $\tilde{p}_i(t)$ is distributed to nearby grid cells using truncated Gaussian splatting with bandwidth $\sigma_P=1$ grid cell and support radius $r_P=2$ grid cells. The generation value at grid cell $g$ is obtained by normalized weighted aggregation:
\begin{equation}
P_t(g)
=
\frac{
\sum_{i\in\mathcal{S}}
w_{ig}\tilde{p}_i(t)
}{
\sum_{i\in\mathcal{S}}
w_{ig}+\epsilon
},
\label{eq:pv_field_construction}
\end{equation}
where $w_{ig}$ denotes the truncated Gaussian weight between source station $i$ and grid cell $g$, and $\epsilon$ is a small constant for numerical stability.

To account for the nonuniform spatial coverage of source stations, the support mask $\mathbf{M}_t$ identifies grid cells with sufficient observational support. Cells whose accumulated Gaussian weight exceeds $10^{-6}$ are marked as valid. The mask is provided to the field forecaster, while unsupported cells are excluded from field-level supervision and evaluation. Together, $(\mathbf{P}_t,\mathbf{M}_t)$ defines the source-supported regional PV generation state used in subsequent forecasting.

\subsection{Future Regional PV Generation Field Forecasting}
\label{sec:field_forecasting}

Representing distributed PV measurements as spatially structured fields enables the spatiotemporal evolution of regional generation patterns to be modeled explicitly. However, their limited geographical coverage makes it difficult to capture the influence of approaching cloud systems outside the observed region, particularly near the field boundaries. Satellite observations provide a broader spatial view of cloud and radiation conditions, offering complementary information on atmospheric evolution. We therefore first predict future shortwave radiation (SWR) fields from historical satellite observations over a larger surrounding region. The resulting irradiance forecasts are then combined with historical PV generation fields and future meteorological information to predict the evolution of regional PV generation.

\subsubsection{Satellite-Derived Future Irradiance Guidance}
\label{sec:swr_guidance}

Historical satellite observations are collected over a broader spatial domain spanning \(26^\circ\)--\(34^\circ\)N and \(116^\circ\)--\(126.5^\circ\)E, covering the Ningbo PV region and surrounding inland, coastal, and offshore areas.

Given four hourly multichannel satellite observations from \(T-4\)~h to \(T-1\)~h, the irradiance forecasting model predicts three hourly-mean SWR fields labeled at \(T\), \(T+1\)~h, and \(T+2\)~h, as illustrated in Fig.~\ref{fig:framework}. The latest observed SWR field over the target region is also retained as historical context for subsequent temporal alignment.

To reduce the influence of the diurnal solar cycle, irradiance is represented using a clear-sky-normalized ratio:
\begin{equation}
Q_t(x,y)=
\begin{cases}
\operatorname{clip}\left(
\dfrac{R_t(x,y)}{R_t^{\mathrm{cs}}(x,y)},
0,Q_{\max}
\right),
& R_t^{\mathrm{cs}}(x,y)\ge R_{\min}^{\mathrm{cs}},\\[6pt]
0,
& R_t^{\mathrm{cs}}(x,y)<R_{\min}^{\mathrm{cs}},
\end{cases}
\label{eq:swr_ratio}
\end{equation}
where $\mathbf{R}_t$ and $\mathbf{R}_t^{\mathrm{cs}}$ denote the historical SWR and corresponding clear-sky irradiance fields, respectively. The clear-sky irradiance threshold is $R_{\min}^{\mathrm{cs}}=50~\mathrm{W\,m^{-2}}$, and the ratio is clipped at $Q_{\max}=1.6$.

The predicted SWR fields are reconstructed from the predicted ratios and clear-sky irradiance:
\begin{equation}
\widehat{\mathbf{R}}_t
=
\operatorname{clip}\left(
\widehat{\mathbf{Q}}_t\odot\mathbf{R}_t^{\mathrm{cs}},
0,R_{\max}
\right),
\end{equation}
where $R_{\max}=1200~\mathrm{W\,m^{-2}}$. The historical ratio inputs and future irradiance guidance used for PV-field forecasting are set to zero where clear-sky irradiance falls below $R_{\min}^{\mathrm{cs}}$.

The irradiance forecasting model consists of a 3D spatiotemporal base predictor followed by a residual ConvLSTM correction stage. The predicted SWR fields are cropped to the PV region and used to guide subsequent regional generation-field forecasting.

\subsubsection{Temporal Alignment of Irradiance Guidance}
\label{sec:swr_alignment}

The irradiance guidance is available at an hourly resolution, whereas the regional PV generation field is forecast at 15-min intervals. A fixed temporal alignment is therefore applied to obtain horizon-specific irradiance guidance for each PV-field forecast step.

The anchor sequence consists of the latest observed irradiance guidance and three predicted future irradiance guidance representations:
\begin{equation}
\left[
\mathbf{A}_0,
\mathbf{A}_1,
\mathbf{A}_2,
\mathbf{A}_3
\right]
=
\left[
\mathbf{R}^{\mathrm{obs}}_{T-1\mathrm{h}},
\widehat{\mathbf{R}}_{T},
\widehat{\mathbf{R}}_{T+1\mathrm{h}},
\widehat{\mathbf{R}}_{T+2\mathrm{h}}
\right].
\label{eq:swr_anchors}
\end{equation}

As these SWR fields represent hourly means, temporal interpolation is performed using the centers of their corresponding averaging intervals. The four anchor centers are therefore located at (T-30), (T+30), (T+90), and (T+150) min, respectively.

For the eight 15-min PV-field forecast timestamps
\(\tau_n=T+n\Delta t\), where \(\Delta t=15\) min and
\(n=0,\ldots,7\), the aligned SWR sequence is obtained by
\begin{equation}
\begin{bmatrix}
\widetilde{\mathbf{R}}_{\tau_0}\\
\widetilde{\mathbf{R}}_{\tau_1}\\
\widetilde{\mathbf{R}}_{\tau_2}\\
\widetilde{\mathbf{R}}_{\tau_3}\\
\widetilde{\mathbf{R}}_{\tau_4}\\
\widetilde{\mathbf{R}}_{\tau_5}\\
\widetilde{\mathbf{R}}_{\tau_6}\\
\widetilde{\mathbf{R}}_{\tau_7}
\end{bmatrix}
=
\mathbf{W}
\begin{bmatrix}
\mathbf{A}_0\\
\mathbf{A}_1\\
\mathbf{A}_2\\
\mathbf{A}_3
\end{bmatrix},
\label{eq:swr_alignment}
\end{equation}
where
\begin{equation}
\mathbf{W}
=
\begin{bmatrix}
0.50 & 0.50 & 0    & 0\\
0.25 & 0.75 & 0    & 0\\
0    & 1.00 & 0    & 0\\
0    & 0.75 & 0.25 & 0\\
0    & 0.50 & 0.50 & 0\\
0    & 0.25 & 0.75 & 0\\
0    & 0    & 1.00 & 0\\
0    & 0    & 0.75 & 0.25
\end{bmatrix}.
\label{eq:swr_weight_matrix}
\end{equation}

The weights are fixed according to linear interpolation between neighboring anchor centers and introduce no trainable parameters. This alignment converts the hourly SWR guidance into a 15-min sequence consistent with the temporal resolution of regional PV-field forecasting.

\subsubsection{Spatiotemporal Field Evolution Modeling}
\label{sec:field_evolution}

The field forecasting model integrates three sources of information: historical PV generation fields, temporally aligned future irradiance guidance, and future meteorological and solar-condition features, as illustrated in Fig.~\ref{fig:framework}. Historical PV fields capture recent regional generation dynamics, while future irradiance and meteorological information provide guidance on upcoming atmospheric conditions. A 3D convolutional encoder extracts spatiotemporal features from historical PV fields and their auxiliary inputs, while irradiance information is processed through separate convolutional encoders. These features are combined with horizon-specific future conditions to predict the evolution of regional PV generation.

Regional generation patterns can evolve through both spatial propagation and local intensity changes. To capture these complementary dynamics, the prediction head consists of two branches. The spatial-propagation branch estimates a horizon-specific displacement field and warps the latest observed PV field to model the movement of existing generation patterns. In parallel, the direct-prediction branch estimates future generation fields without imposing spatial displacement constraints, capturing local intensity changes that cannot be represented by warping alone.

The predicted displacement is bounded to three grid cells in each spatial direction, and warping is performed using differentiable bilinear sampling with border padding. A pixel-wise fusion gate is generated from the shared forecasting features and the outputs of both prediction branches. The two predictions are then adaptively combined at each grid cell:
\begin{equation}
\widehat{\mathbf{P}}_{\tau_n}
=
\mathbf{G}_{n}
\odot
\widehat{\mathbf{P}}^{\mathrm{warp}}_{\tau_n}
+
\left(1-\mathbf{G}_{n}\right)
\odot
\widehat{\mathbf{P}}^{\mathrm{dir}}_{\tau_n},
\label{eq:motion_fusion}
\end{equation}
where $\mathbf{G}_{n}\in[0,1]^{H_g\times W_g}$ denotes the learned pixel-wise fusion gate at forecast horizon $\tau_n$. The resulting sequence
$\{\widehat{\mathbf{P}}_{\tau_n}\}_{n=0}^{H-1}$
represents the predicted future regional PV generation fields.

The field forecaster is optimized using a support-masked SmoothL1 reconstruction loss:
\begin{equation}
\mathcal{L}_{\mathrm{rec}}
=
\operatorname{SmoothL1}
\left(
\widehat{\mathbf{P}}_{\tau_n},
\mathbf{P}_{\tau_n}
\right),
\label{eq:field_loss}
\end{equation}
where the loss is averaged over source-supported grid cells and forecast horizons.

\subsection{Regional-to-Local Station Forecasting}
\label{sec:regional_to_local}

The predicted regional PV generation fields provide future spatial context for individual target stations. For each target station, location-specific field patches are extracted according to its geographical coordinates and combined with its historical observations and future meteorological and solar-condition information for station-level forecasting.

For target station $j$ and forecast timestamp $\tau_n$, a $5\times5$ local patch is extracted from the corresponding predicted regional field:
\begin{equation}
\begin{aligned}
\mathbf{X}^{\mathrm{field}}_{j,\tau_n}
=
\mathcal{C}_j\big(
&\widehat{\mathbf{P}}_{\tau_n},\\
&\mathbf{M}_{T-\Delta t}
\big),
\end{aligned}
\label{eq:field_patch}
\end{equation}
where $\mathcal{C}_j(\cdot)$ denotes location-based cropping centered at station $j$. Each patch contains two channels: the predicted PV generation field and the source-support mask. The latest observed support mask is reused across all forecast horizons to identify source-supported grid cells. The predicted-field channel is standardized using statistics computed from the training-set patches, while the support mask remains binary.

The station-level forecast integrates three sources of information: the target station's historical observations, future meteorological and solar-condition features, and the predicted regional field context:
\begin{equation}
\begin{aligned}
\widehat{\mathbf{p}}^{\,\mathrm{pu}}_{j,T:T+H-1}
=
f_{\mathrm{station}}\big(
&\mathbf{X}^{\mathrm{hist}}_{j,T},\\
&\mathbf{Z}^{\mathrm{fut}}_{j,T},\\
&\mathbf{X}^{\mathrm{field,fut}}_{j,T}
\big),
\end{aligned}
\label{eq:station_forecasting}
\end{equation}
where $\mathbf{X}^{\mathrm{hist}}_{j,T}$ contains historical PV observations and associated station-level features, $\mathbf{Z}^{\mathrm{fut}}_{j,T}$ contains future meteorological and solar-condition information, and $\mathbf{X}^{\mathrm{field,fut}}_{j,T}$ denotes the sequence of location-specific patches extracted from the predicted regional fields.

The historical station sequence is encoded using a bidirectional LSTM, while future-condition features and regional field patches are separately encoded using multilayer perceptrons with weights shared across forecast horizons. A learnable horizon embedding distinguishes individual forecast steps. These representations are fused to produce horizon-specific station-level PV forecasts.

The station forecaster is trained using a weighted SmoothL1 loss applied to station-specific standardized PV outputs. Forecast horizons with a ground-truth power change of at least $0.10$~p.u. relative to the latest historical observation receive a weight of $1.5$, while the remaining horizons receive unit weight. The weighted loss is normalized by the sum of the sample weights.

\section{Experiments}

\subsection{Experimental Setup and Evaluation Protocol}
\label{sec:experimental_setup}

The proposed framework is evaluated on a real-world distributed PV dataset collected from 496 PV systems in Ningbo, Zhejiang Province, China. The dataset spans January 1, 2022 to June 24, 2023 at a 15-min resolution. The forecasting task uses twelve historical observations covering the preceding three hours to predict the subsequent eight time steps, corresponding to forecast horizons of 15--120~min.

Among the 496 systems, 446 stations form the source fleet for regional field construction and supervision, while 50 spatially distributed stations serve as target sites for downstream forecasting. The target stations are excluded from regional field construction and supervision, but their historical measurements and training-period labels remain available for station-level model training. Performance is evaluated at both regional-field and station levels.

\subsubsection{Dataset and Auxiliary Data}

Only daytime observations from 07:00 to 16:45 Beijing time are retained. PV power is normalized by installed capacity, while the additional station-wise scaling used for regional PV-field construction follows Section~\ref{sec:field_construction}.

Hourly Himawari-8 Level-3 satellite products are obtained from the JAXA P-Tree system \cite{jaxa_ptree_himawari,bessho2016introduction}. The satellite inputs include shortwave radiation (SWR), photosynthetically active radiation (PAR), ultraviolet-A (UVA), ultraviolet-B (UVB), aerosol optical thickness (AOT), and Ångström exponent (AE). Historical meteorological data are obtained from ERA5 \cite{hersbach2020era5}, while future meteorological information is provided by a commercial operational weather-forecast product issued at 06:00 BJT on the preceding day. The forecasts include radiation, 10-m wind direction and speed, and total cloud cover at 12 representative locations across the PV region. These variables are interpolated onto a $16\times16$ grid using inverse-distance weighting and subsequently resized to the $64\times64$ PV-field grid. Location-specific weather forecasts are used for station-level prediction. Solar-position variables, clear-sky GHI, and cyclical time encodings are also included as auxiliary features.

\begin{figure}[!t]
\centering
\includegraphics[width=\columnwidth]{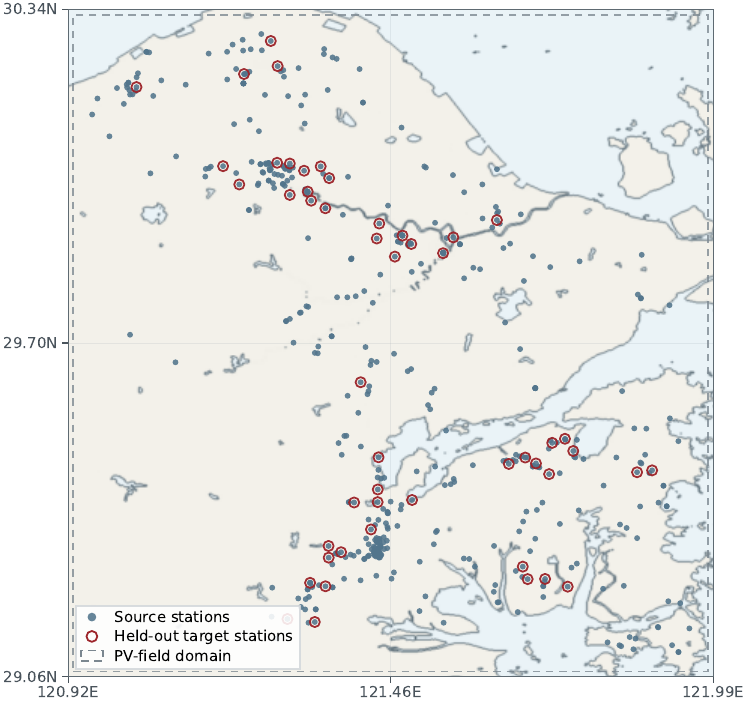}
\caption{Spatial distribution of 446 source stations and 50 held-out target stations in the PV-field domain.}
\label{fig:station_distribution}
\end{figure}

\subsubsection{Source--Target Setting and Data Split}

As shown in Fig.~\ref{fig:station_distribution}, 446 stations are used as the source fleet and 50 stations are retained as target sites. Target candidates are required to have at least eight neighboring PV stations within 8~km. The final targets are selected using Latin-hypercube-based spatial stratification with a minimum separation of 2~km, while each selected target retains at least five non-target source stations within 5~km. Target stations are excluded from regional PV-field construction and field-level supervision throughout the experiments.

The data are chronologically divided into training, validation, and test sets. The training period spans January 1, 2022 to January 13, 2023, the validation period spans January 14 to March 8, 2023, and the test period spans March 9 to June 24, 2023. All normalization and standardization statistics are estimated from the training period and then fixed for validation and testing.

\subsubsection{Forecasting Protocol}

For each forecast issue time \(T\), the latest historical PV observation is available at \(T-15\)~min. Twelve historical observations covering the preceding three hours are used to predict eight future steps at 15-min resolution, corresponding to horizons of 15--120~min ahead of the latest observation. Six forecast origins are used each day at 10:00, 11:00, 12:00, 13:00, 14:00, and 15:00 BJT. The same forecast origins are used for regional PV-field and station-level forecasting.

Historical PV and station-level meteorological observations are available up to \(T-15\)~min. For satellite-derived irradiance guidance, the latest observed Himawari-8 hourly field is labeled at \(T-1\)~h and represents the averaging interval \([T-1\mathrm{h},T]\). Future meteorological inputs are obtained from forecasts issued at 06:00 BJT on the preceding day. Satellite inputs are restricted to observations covering periods no later than the forecast issue time. The evaluation assumes that these observations are available at their corresponding observation times, without accounting for satellite product processing, transmission, or dissemination delays.

This protocol yields 2268 training, 324 validation, and 648 test forecast-origin instances. The three forecasting stages are trained sequentially. Once trained, the irradiance and regional field forecasters are fixed, and their predicted outputs are used to train the downstream station-level forecaster.

\subsubsection{Evaluation Metrics}

Regional PV-field forecasting is evaluated using MAE, RMSE, $R^2$, masked global SSIM, DeltaCorr, SignAcc, and Large-$\Delta$ MAE. Station-level forecasting is evaluated using MAE, RMSE, and $R^2$. Regional-field errors are evaluated on the normalized generation fields defined in Section~\ref{sec:field_construction}, while station-level errors are reported in per-unit power normalized by installed capacity. Regional-field metrics are computed over grid cells supported by the source-station mask.

For SSIM, global means, variances, and covariance are computed over source-supported pixels within each predicted and observed field pair, rather than using sliding local windows. The resulting frame-level SSIM scores are averaged equally across valid forecast origins and horizons.

To characterize the evolution of the regional generation state, the predicted and observed increments relative to the latest observed field are defined as
\begin{equation}
\Delta\widehat{\mathbf{P}}_{s,h}
=
\widehat{\mathbf{P}}_{s,h}
-
\mathbf{P}^{\mathrm{last}}_{s},
\qquad
\Delta\mathbf{P}_{s,h}
=
\mathbf{P}_{s,h}
-
\mathbf{P}^{\mathrm{last}}_{s},
\label{eq:field_increment_metric}
\end{equation}
where $s$ denotes a forecast-origin sample and $h$ denotes a forecast horizon.

DeltaCorr is the Pearson correlation between the predicted and observed increments over supported pixels, computed for each valid sample--horizon frame and then averaged across frames. Frames with zero variance in either increment field are excluded. SignAcc measures the fraction of supported pixels for which the predicted and observed increments have the same sign, excluding negligible observed changes with $|\Delta P|\leq10^{-6}$. Large-$\Delta$ MAE evaluates field-value errors over supported pixels with substantial observed changes, defined by $|\Delta P|\geq0.10$.

Because different methods yield slightly different sets of valid test samples, all methods within each comparison are evaluated on their common valid subset to ensure a fair comparison. For the main station-level comparisons, this results in 623 valid forecast origins out of the original 648. Station-wise MAE is also computed for each target station, and the win count indicates the number of stations where a method achieves lower MAE than the corresponding reference method.

\subsection{Validation of Regional PV Generation Field Representation}

To examine how the constructed PV generation fields represent regional generation patterns under different conditions, we select three representative operating regimes: stable high-generation, spatially heterogeneous generation, and rapidly varying generation. These regimes are identified using source-fleet statistics, including regional mean generation, spatial standard deviation, and 15-min generation changes, which characterize overall generation level, spatial heterogeneity, and short-term variability, respectively.

\begin{figure}[t]
    \centering
    \includegraphics[width=0.46\textwidth]{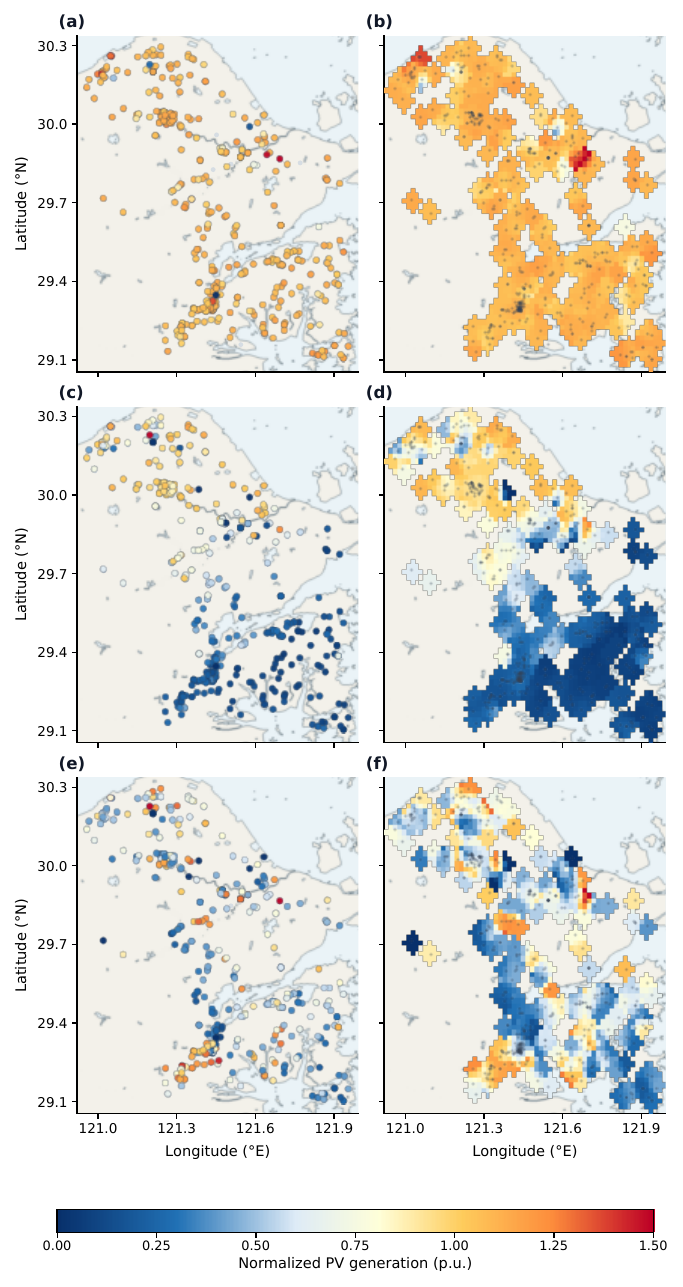}
    \caption{
    Representative source-fleet PV observations and their corresponding regional PV generation fields under different generation regimes.
    Panels (a)--(b), (c)--(d), and (e)--(f) correspond to stable high-generation, spatially heterogeneous generation, and rapidly varying generation, respectively.
    The left column shows normalized PV measurements at source stations, while the right column shows the constructed source-supported regional PV generation fields.
    }
    \label{fig:spatial_representation}
\end{figure}

Fig.~\ref{fig:spatial_representation} compares discrete source-station measurements with the corresponding regional PV generation fields under the three operating regimes. The stable high-generation case exhibits relatively uniform generation across the region ($\mu=1.080$, $\sigma=0.134$, and $\Delta=0.012$), while the spatially heterogeneous case shows more pronounced regional differences ($\sigma=0.355$). The rapidly varying case exhibits both substantial spatial heterogeneity and the largest 15-min generation change ($\Delta=0.230$). Across these conditions, the constructed fields make the spatial distribution and variation of regional PV generation explicit, providing a structured representation of the underlying source-fleet observations.

To evaluate the consistency between the constructed fields and the original source-station measurements, we sampled the fields at source-station locations and compared the resulting values with the corresponding normalized observations. Across 1,907,857 station-time samples, the MAE, RMSE, and $R^2$ were 0.0539, 0.1037, and 0.9025, respectively. These results indicate that the constructed fields retain the main generation characteristics of the source fleet while providing a spatially structured regional representation.

\subsection{Forecasting Performance of Regional PV Generation Fields}
\label{sec:field_results}

We evaluate the forecasting performance of regional PV generation fields over the 15--120~min horizons. The proposed field forecaster is compared with persistence and four U-Net baselines using different combinations of historical PV fields, future meteorological forecasts, and historical irradiance information. Performance is assessed in terms of field-value accuracy, global structural similarity, and the consistency of predicted generation changes.

\begin{table*}[!t]
\centering
\caption{Regional PV-field forecasting performance. Large-$\Delta$ MAE is evaluated over supported pixels with substantial observed changes ($|\Delta P|\geq0.10$).}
\label{tab:pv_field_main}
\footnotesize
\setlength{\tabcolsep}{5.5pt}
\renewcommand{\arraystretch}{1.08}
\begin{tabular}{lccccccc}
\hline
\textbf{Method}
& \textbf{MAE} $\downarrow$
& \textbf{RMSE} $\downarrow$
& \textbf{$R^2$} $\uparrow$
& \textbf{SSIM} $\uparrow$
& \textbf{DeltaCorr} $\uparrow$
& \textbf{SignAcc} $\uparrow$
& \textbf{Large-$\Delta$ MAE} $\downarrow$ \\
\hline
Persistence
& 0.1483
& 0.2077
& 0.6098
& 0.5516
& --
& 0.0000
& 0.2520 \\

PV-only U-Net
& 0.0957
& 0.1398
& 0.8233
& 0.6106
& 0.5760
& 0.7677
& 0.1295 \\

PV+NWP U-Net
& 0.0942
& 0.1374
& 0.8293
& 0.6043
& 0.5895
& 0.7736
& 0.1255 \\

PV+Past SWR U-Net
& 0.0983
& 0.1428
& 0.8157
& 0.5998
& 0.5576
& 0.7571
& 0.1334 \\

PV+Past SWR+NWP U-Net
& 0.0950
& 0.1392
& 0.8247
& 0.6004
& 0.5737
& 0.7706
& 0.1281 \\

Proposed field forecaster
& \textbf{0.0880}
& \textbf{0.1306}
& \textbf{0.8456}
& \textbf{0.6297}
& \textbf{0.6132}
& \textbf{0.7773}
& \textbf{0.1207} \\
\hline
\end{tabular}
\end{table*}

Table~\ref{tab:pv_field_main} summarizes the regional field forecasting results. Persistence produces relatively large errors, with an MAE of 0.1483 and an RMSE of 0.2077. The U-Net baselines substantially reduce these errors, achieving MAEs between 0.0942 and 0.0983. Among them, PV+NWP U-Net provides the best point-wise accuracy, with an MAE of 0.0942 and an $R^2$ of 0.8293.

The proposed field forecaster achieves the best performance across all reported metrics, reducing MAE to 0.0880 and RMSE to 0.1306. Compared with PV+NWP U-Net, this corresponds to a 6.6\% reduction in MAE. The proposed model also achieves the highest SSIM of 0.6297 and DeltaCorr of 0.6132, while reducing Large-$\Delta$ MAE to 0.1207. These results demonstrate improved accuracy in predicting regional generation fields, including their overall spatial patterns and temporal changes.

\begin{figure}[!t]
\centering
\includegraphics[width=0.5\textwidth]{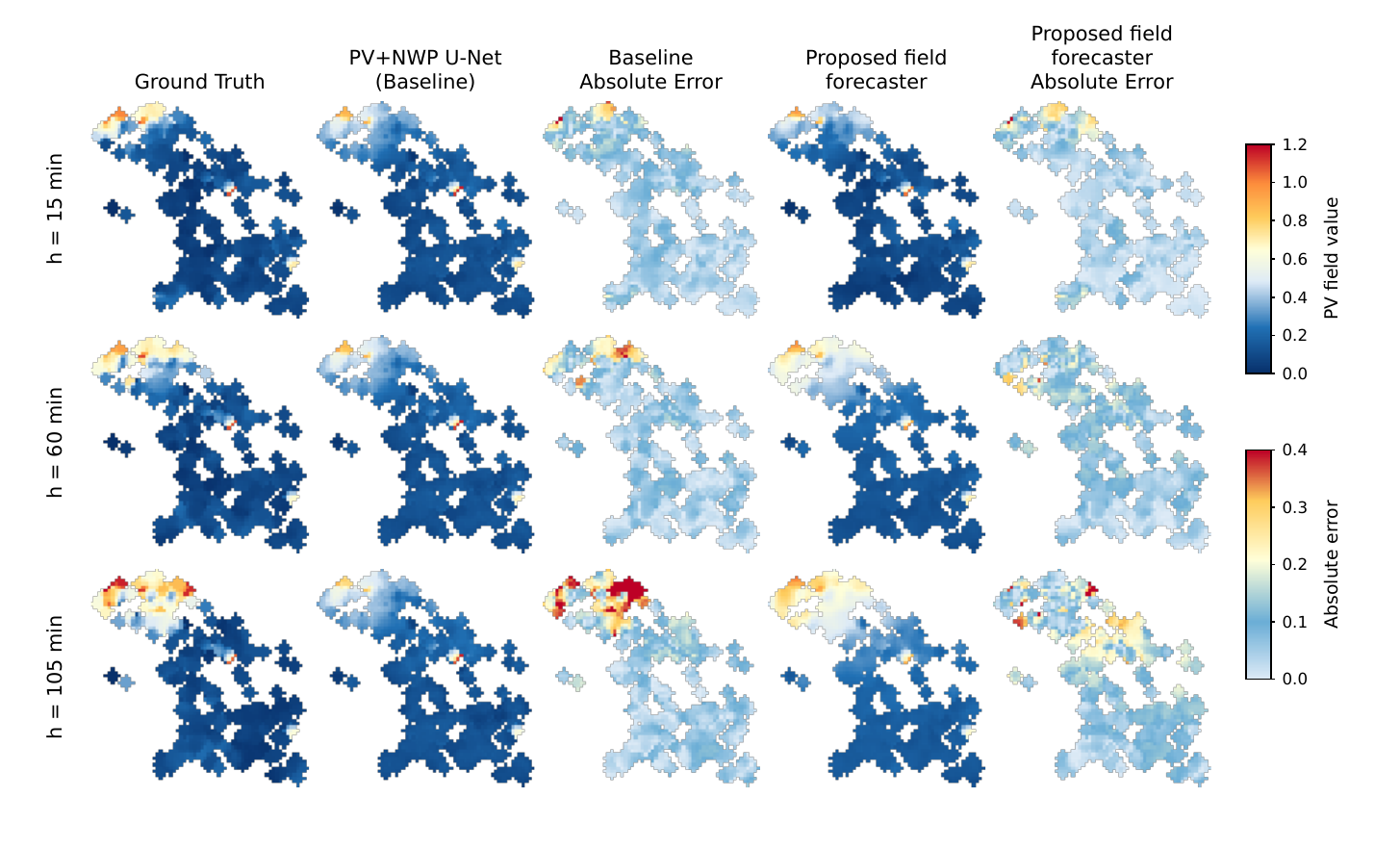}
\caption{Qualitative comparison of regional PV-field forecasts at three forecast lead times. Columns show the ground truth, the PV+NWP U-Net prediction and absolute error, and the proposed field prediction and absolute error. All field maps share the same value scale, while both error maps share the same error scale.}
\label{fig:field_examples}
\end{figure}


Fig.~\ref{fig:field_examples} compares regional field forecasts at 15-, 60-, and 105-min lead times. At 15~min, both methods reproduce the dominant spatial generation pattern. As the forecast horizon increases, the differences become more apparent. At 105~min, the proposed model more accurately captures the high-generation region in the northern part of the field, with smaller and less spatially extensive prediction errors. This example illustrates the improvement in spatial field prediction observed in the quantitative evaluation.

\begin{figure}[!t]
\centering
\includegraphics[width=\columnwidth]{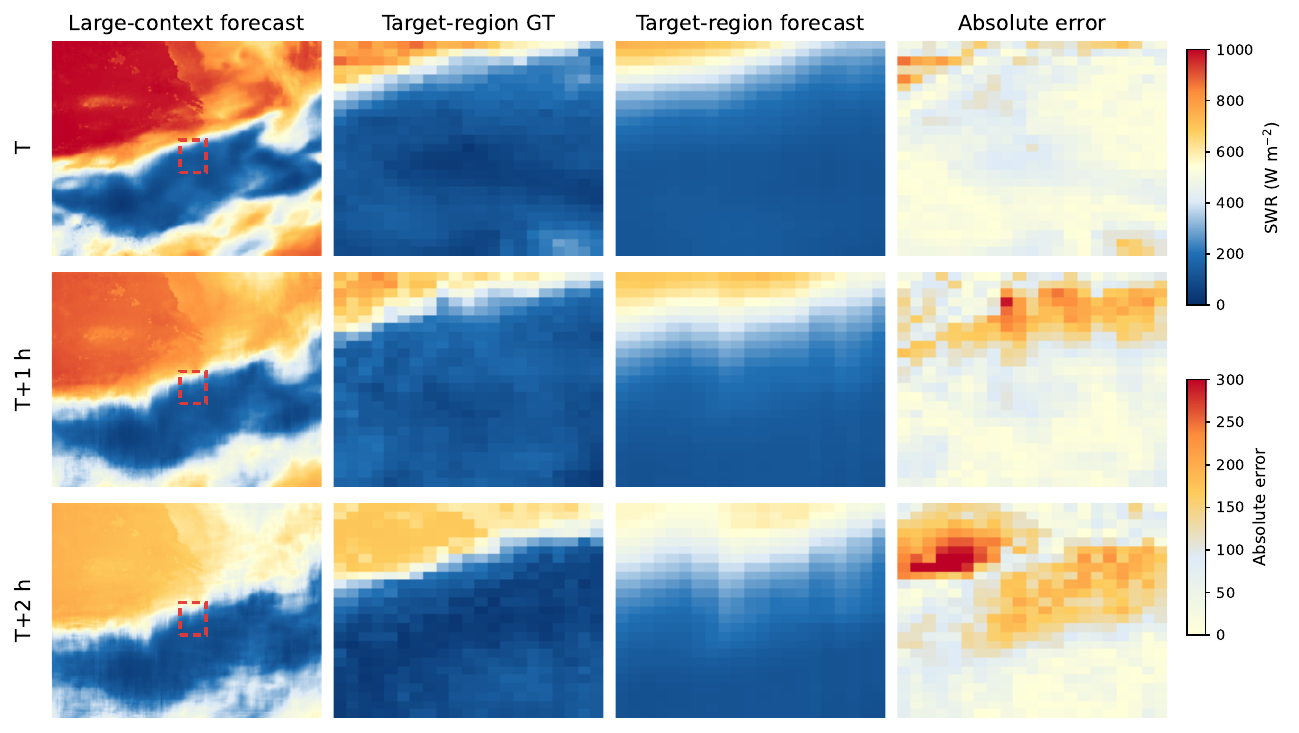}
\caption{Example of satellite-derived future irradiance guidance. The red dashed box marks the Ningbo PV region within each large-context forecast. The extracted target-region forecasts are shown with the corresponding ground-truth SWR fields and absolute errors at the three future guidance anchors.}
\label{fig:irradiance_guidance}
\end{figure}

Fig.~\ref{fig:irradiance_guidance} presents an example of satellite-derived future irradiance guidance. The large-context SWR forecasts describe the evolving irradiance distribution over the surrounding region, while the extracted target-region fields are compared with the corresponding observations. These forecasts are subsequently aligned with the PV-field prediction timestamps to provide horizon-specific irradiance guidance.

\begin{table}[!t]
\centering
\caption{Ablation study of future irradiance guidance and regional field-evolution modeling.}
\label{tab:field_ablation}
\scriptsize
\setlength{\tabcolsep}{4.5pt}
\renewcommand{\arraystretch}{1.05}
\begin{tabular}{lcccc}
\hline
\textbf{Setting} &
\textbf{MAE}$\downarrow$ &
\textbf{RMSE}$\downarrow$ &
\textbf{$R^2$}$\uparrow$ &
\textbf{SSIM}$\uparrow$ \\
\hline
w/o future irradiance guidance
& 0.0898 & 0.1336 & 0.8385 & 0.6032 \\

w/o temporal alignment
& 0.0911 & 0.1350 & 0.8352 & 0.6231 \\

Direct-only head
& 0.0935 & 0.1336 & 0.8386 & 0.6107 \\

Warp-only head
& 0.1029 & 0.1445 & 0.8111 & 0.5614 \\

w/o adaptive gate
& 0.0992 & 0.1389 & 0.8255 & 0.5822 \\

Full model
& \textbf{0.0880}
& \textbf{0.1306}
& \textbf{0.8456}
& \textbf{0.6297} \\
\hline
\end{tabular}
\end{table}

Table~\ref{tab:field_ablation} examines the contributions of future irradiance guidance, temporal alignment, and the field-evolution prediction head. Removing future irradiance guidance increases MAE from 0.0880 to 0.0898 and reduces SSIM from 0.6297 to 0.6032. Without temporal alignment, MAE increases to 0.0911. These results indicate that both future irradiance information and its temporal alignment contribute to regional field forecasting accuracy.

The field-evolution ablations show more pronounced differences. The direct-only model achieves an MAE of 0.0935, while the warp-only model produces a higher MAE of 0.1029 and a lower SSIM of 0.5614. Removing the adaptive gate also degrades performance, increasing MAE to 0.0992. The full model consistently outperforms these variants, supporting the complementary roles of spatial propagation and local intensity changes, as well as the benefit of their adaptive fusion.

\subsection{Benefits of Regional Field Context for Station-Level Forecasting}
\label{sec:regional_context}

We investigate whether predicted regional PV generation fields provide useful information for target-station forecasting beyond local observations and meteorological forecasts. Different sources and representations of regional information are compared using a common BiLSTM-based station forecasting backbone. All variants use the same target-site historical observations and future meteorological and solar-condition information, differing only in their regional guidance inputs and associated context encoders.

The local-only BiLSTM+NWP model serves as the baseline without regional guidance. The 3-NN PV variant incorporates historical generation measurements from the three nearest source stations, while future irradiance guidance provides predicted spatial irradiance information. We also compare historical PV-field guidance, extracted from the latest observed regional field, with predicted future PV-field guidance to assess the benefit of forecasting regional generation evolution.

A graph-based spatiotemporal approach is further included to compare different representations of regional generation dynamics. A fixed geographical 8-nearest-neighbor graph is constructed from the same 446 source stations, excluding all 50 target stations. A graph-based GRU predicts future source-node generation states using historical PV measurements and future meteorological and irradiance information. For each target station, source nodes within its local field region are aggregated through distance-weighted pooling to provide location-specific context. Historical graph guidance uses the latest observed node states, while predicted graph guidance uses their forecasted future states. Both variants share the same source fleet and station-level forecasting backbone as the corresponding PV-field approaches.

\begin{table}[!t]
\centering
\caption{Comparison of regional information sources under a common station-level forecasting backbone. All methods are evaluated on the same test records.}
\label{tab:regional_guidance}
\scriptsize
\setlength{\tabcolsep}{4.5pt}
\renewcommand{\arraystretch}{1.08}
\resizebox{\columnwidth}{!}{
\begin{tabular}{lccc}
\hline
\textbf{Regional information}
& \textbf{MAE} $\downarrow$
& \textbf{RMSE} $\downarrow$
& \textbf{$R^2$} $\uparrow$ \\
\hline

None (BiLSTM+NWP)
& 0.0746 & 0.1084 & 0.7956 \\

3-NN PV guidance
& 0.0737 & 0.1074 & 0.7995 \\

Future irradiance guidance
& 0.0713 & 0.1033 & 0.8144 \\
\hline

Historical PV-field guidance
& 0.0725 & 0.1060 & 0.8048 \\

Predicted future PV-field guidance
& \textbf{0.0672}
& \textbf{0.0999}
& \textbf{0.8266} \\
\hline

Historical graph guidance
& 0.0719 & 0.1058 & 0.8054 \\

Predicted graph guidance
& 0.0689 & 0.1019 & 0.8197 \\
\hline
\end{tabular}}
\end{table}

Table~\ref{tab:regional_guidance} shows that all forms of regional guidance improve station-level forecasting over the local-only baseline. The 3-NN PV variant achieves a modest reduction in MAE from 0.0746 to 0.0737, while future irradiance guidance further reduces it to 0.0713. Historical PV-field guidance also improves performance, achieving an MAE of 0.0725. More importantly, using predicted future PV fields reduces MAE to 0.0672, representing a 7.3\% improvement over historical PV-field guidance and a 5.8\% improvement over future irradiance guidance. These results demonstrate the additional value of explicitly forecasting future regional PV generation states.

The graph-based results provide a complementary comparison. Predicting future graph-node states reduces MAE from 0.0719 to 0.0689 compared with historical graph guidance, confirming the benefit of forecasting future fleet-level generation dynamics. The predicted PV-field approach further reduces MAE by 2.5\% relative to predicted graph guidance and achieves lower station-wise MAE at 40 of the 50 target stations. This indicates that the explicit regional field representation provides effective spatial context for downstream station forecasting.

Overall, predicted future PV-field guidance achieves the best performance, reducing station-level MAE by approximately 10.0\% relative to the local-only baseline. The results highlight the benefit of predicting regional generation dynamics and transferring the resulting spatial context to individual stations.

\subsection{Comparison with Existing Forecasting Methods}
\label{sec:method_comparison}

We compare SatPV-Field with representative temporal and spatiotemporal forecasting methods on the 50 target stations. All methods follow the same chronological data split and forecasting protocol, with access to future meteorological and solar-condition information. The temporal baselines include LSTM, BiLSTM, TCN, Transformer, and PatchTST, while STGCN and AGCRN are included as graph-based spatiotemporal baselines that jointly model the 50 target stations.

\begin{table}[!t]
\centering
\caption{Station-level forecasting performance compared with existing temporal and spatio-temporal methods.}
\label{tab:method_comparison}
\scriptsize
\setlength{\tabcolsep}{3.5pt}
\renewcommand{\arraystretch}{1.08}
\resizebox{\columnwidth}{!}{
\begin{tabular}{llccc}
\hline
\textbf{Type}
& \textbf{Method}
& \textbf{MAE} $\downarrow$
& \textbf{RMSE} $\downarrow$
& \textbf{$R^2$} $\uparrow$ \\
\hline

Temporal
& LSTM+NWP
& 0.0754 & 0.1092 & 0.7927 \\

Temporal
& BiLSTM+NWP
& 0.0746 & 0.1084 & 0.7956 \\

Temporal
& TCN+NWP
& 0.0753 & 0.1086 & 0.7949 \\

Temporal
& Transformer+NWP
& 0.0783 & 0.1125 & 0.7799 \\

Temporal
& PatchTST+NWP
& 0.0778 & 0.1130 & 0.7780 \\
\hline

Spatio-temporal
& STGCN+NWP
& 0.0758 & 0.1082 & 0.7965 \\

Spatio-temporal
& AGCRN+NWP
& 0.0694 & 0.1035 & 0.8137 \\
\hline

Proposed
& SatPV-Field
& \textbf{0.0672}
& \textbf{0.0999}
& \textbf{0.8266} \\
\hline
\end{tabular}}
\end{table}

Table~\ref{tab:method_comparison} summarizes the station-level forecasting results. Among the temporal baselines, BiLSTM+NWP achieves the lowest MAE of 0.0746. The graph-based models show different performance levels: STGCN remains comparable to the temporal baselines, while AGCRN achieves a lower MAE of 0.0694, demonstrating the potential benefit of modeling cross-station dependencies.

SatPV-Field achieves the best performance across all three metrics, with an MAE of 0.0672, an RMSE of 0.0999, and an $R^2$ of 0.8266. Compared with BiLSTM+NWP and AGCRN+NWP, it reduces MAE by 9.9\% and 3.2\%, respectively. These results demonstrate the effectiveness of the proposed regional-to-local forecasting framework relative to both temporal and graph-based spatiotemporal approaches.

\subsection{Robustness of Regional Field Forecasting and Guidance}
\label{sec:robustness}

We examine the performance of regional PV-field forecasting and its downstream station-level guidance across forecast horizons, source-fleet densities, and different operating conditions.

\subsubsection{Performance Across Forecast Horizons}

\begin{figure}[t]
    \centering
    \subfloat[]{%
        \includegraphics[width=0.48\columnwidth]{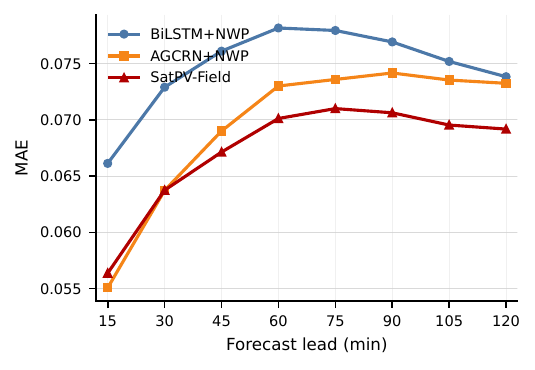}
        \label{fig:horizon_station_mae}
    }\hfill
    \subfloat[]{%
        \includegraphics[width=0.48\columnwidth]{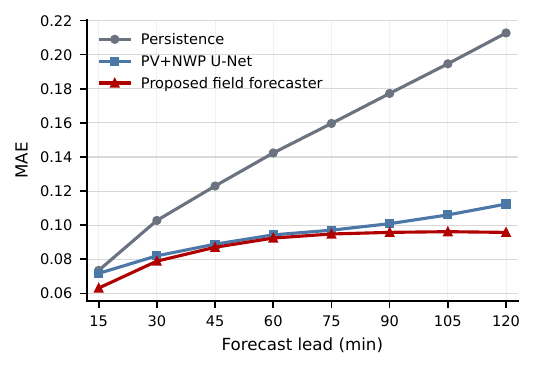}
        \label{fig:horizon_field_mae}
    }
    \caption{Forecasting performance across forecast lead times.
    (a) Station-level MAE for BiLSTM+NWP, AGCRN+NWP, and predicted future PV-field guidance.
    (b) Regional PV-field MAE for Persistence, PV+NWP U-Net, and the proposed field forecaster.}
    \label{fig:horizon_results}
\end{figure}

Fig.~\ref{fig:horizon_results} compares forecasting performance across the eight prediction horizons. At the station level, the local-only BiLSTM+NWP baseline exhibits higher MAE over most horizons. AGCRN+NWP performs competitively at shorter horizons, while predicted future PV-field guidance achieves lower MAE from 45 to 120~min. The advantage becomes more pronounced at medium and longer horizons, highlighting the increasing value of future regional generation context as lead time increases.

At the regional-field level, the proposed forecaster achieves the lowest MAE across all horizons. Persistence errors increase rapidly with lead time, while PV+NWP U-Net shows a more gradual deterioration. In comparison, the proposed model maintains consistently lower errors throughout the 15--120~min forecasting window.

\subsubsection{Source-Fleet Density and Field Observability}

We investigate how source-fleet density affects regional field construction and downstream forecasting. The 446 source stations are reduced to 112, 223, and 335 stations, representing 25\%, 50\%, and 75\% of the full source fleet, respectively. Spatially balanced nested subsets are used, while the same 50 target stations are retained across all settings.

For each density setting, regional fields are reconstructed from the selected source stations, and separate field and station forecasters are trained. Field predictions are evaluated against the full-density regional reference using a common support mask. Local field coverage is measured by the fraction of supported pixels within target-centered patches, while the median distance to the nearest source station characterizes the spatial availability of source observations.

\begin{table}[!t]
\centering
\caption{Sensitivity to source-fleet density. Coverage denotes the supported-pixel ratio within local field patches, and wins denote target stations with lower MAE than BiLSTM+NWP. Station-level results are obtained from independently trained density experiments.}
\label{tab:source_density}
\scriptsize
\setlength{\tabcolsep}{3.5pt}
\renewcommand{\arraystretch}{1.05}
\resizebox{\columnwidth}{!}{
\begin{tabular}{ccccccc}
\hline
\textbf{Ratio} &
\textbf{Sources} &
\textbf{Patch coverage} &
\textbf{Median dist. (km)} &
\textbf{Field MAE} &
\textbf{Station MAE} &
\textbf{Wins/50} \\
\hline
25\%  & 112 & 0.6312 & 2.60 & 0.1251 & 0.0699 & 43 \\
50\%  & 223 & 0.8002 & 1.67 & 0.1063 & 0.0680 & 48 \\
75\%  & 335 & 0.9119 & 1.22 & 0.0914 & 0.0684 & 47 \\
100\% & 446 & 0.9218 & 1.12 & 0.0880 & 0.0670 & 49 \\
\hline
\end{tabular}}
\end{table}

Table~\ref{tab:source_density} shows that increasing source-fleet density improves regional field coverage. The supported-pixel ratio within local patches rises from 0.6312 at 25\% density to 0.9218 at full density, while the median nearest-source distance decreases from 2.60 to 1.12~km. Field MAE correspondingly decreases from 0.1251 to 0.0880, indicating that denser observations improve regional field reconstruction and forecasting.

Station-level performance is less sensitive to source-fleet density. The 50\% and 75\% settings achieve similar MAEs to the full-density configuration, although the improvements are not strictly monotonic. Even at 25\% density, regional field guidance outperforms BiLSTM+NWP at 43 of the 50 target stations. These results indicate that useful regional guidance can be obtained from substantially reduced source fleets, while higher source density improves regional field coverage and accuracy.

\subsubsection{Performance Across Operating Conditions}

We further evaluate station-level forecasting under different cloud-cover and short-term generation-variation conditions. Based on ERA5 total cloud cover (TCC), test samples are classified as clear ($\mathrm{TCC}\leq0.30$), partly cloudy ($0.30<\mathrm{TCC}\leq0.70$), or overcast ($\mathrm{TCC}>0.70$).

Short-term generation variation is measured by the maximum absolute change between consecutive future PV outputs within each target-station forecast window:
\begin{equation}
s_{j,T} =
\max_{n=1,\ldots,H-1}
\left|
p_j^{\mathrm{pu}}(\tau_n)
-
p_j^{\mathrm{pu}}(\tau_{n-1})
\right|,
\label{eq:variation_score}
\end{equation}
where $p_j^{\mathrm{pu}}(t)=p_j(t)/C_j$ denotes the installed-capacity-normalized ground-truth output of target station $j$. Samples are divided into low- and high-variation groups using the median variation score of 0.0853~p.u.

\begin{table}[!t]
\centering
\caption{Station-level forecasting performance under different weather and short-term variation conditions. Reduction is calculated relative to BiLSTM+NWP.}
\label{tab:condition_results}
\scriptsize
\setlength{\tabcolsep}{5.5pt}
\renewcommand{\arraystretch}{1.08}
\begin{tabular}{lccc}
\hline
\textbf{Condition}
& \textbf{BiLSTM+NWP}
& \textbf{PV-field guidance}
& \textbf{Reduction} \\
\hline
\multicolumn{4}{l}{\textit{Weather conditions (MAE)}} \\
Clear
& 0.0472 & \textbf{0.0449} & 4.9\% \\
Partly cloudy
& 0.1004 & \textbf{0.0921} & 8.3\% \\
Overcast
& 0.0801 & \textbf{0.0755} & 5.7\% \\
\hline
\multicolumn{4}{l}{\textit{Short-term variation conditions (RMSE)}} \\
Low variation
& 0.0621 & \textbf{0.0533} & 14.2\% \\
High variation
& 0.1397 & \textbf{0.1304} & 6.7\% \\
\hline
\end{tabular}
\end{table}

Table~\ref{tab:condition_results} shows that predicted regional field guidance improves station-level forecasting across all three cloud-cover categories. The largest relative MAE reduction occurs under partly cloudy conditions (8.3\%), while improvements are also observed under clear and overcast conditions.

Regional field guidance also reduces RMSE in both variation groups, from 0.0621 to 0.0533 under low variation and from 0.1397 to 0.1304 under high variation. Although forecasting errors remain larger during periods of rapid generation changes, the predicted regional context provides consistent improvements under both conditions.

\subsection{Discussion and Practical Implications}
\label{sec:discussion}

The proposed formulation introduces a spatial perspective on distributed PV forecasting by treating fleet-level generation measurements as observations of an evolving regional generation state. Distributed PV output reflects the combined influence of atmospheric conditions and heterogeneous system responses, producing spatial patterns that are not fully characterized by meteorological information alone. Organizing these observations into explicit regional PV generation fields makes such patterns accessible to spatial analysis and future-state prediction. This establishes a regional-to-local forecasting paradigm in which generation dynamics are first modeled at the fleet level and subsequently used to inform individual-site predictions.

The present study provides an initial demonstration of the feasibility of this paradigm. The explicit field representation creates opportunities for spatial and spatiotemporal modeling beyond the architectures explored here. In particular, the current regional-to-local forecasting stage extracts a target-centered $5\times5$ field patch and encodes it using a multilayer perceptron. More advanced approaches, including multiscale convolutional networks, spatial attention, and spatiotemporal vision models, could exploit larger spatial contexts, evolving regional patterns, and interactions across different spatial scales. Such developments may enable richer extraction of generation information from predicted PV fields and further improve downstream forecasting performance.

From a practical perspective, the regional PV field offers a shared spatial representation that connects measurements from existing PV installations with forecasting at other locations. A predicted regional field can provide location-specific generation context without requiring target stations to participate in its construction. Beyond individual-site forecasting, this representation could also support regional generation monitoring, spatial anomaly detection, and distribution-level situational awareness, where geographically organized information on PV generation is valuable.

The current evaluation is limited to a single regional PV fleet, and the reconstructed fields depend on the spatial coverage and characteristics of the available source stations. The forecasting protocol also assumes immediate availability of satellite observations, while operational product latency remains to be considered in real-time deployment. Future research will investigate cross-region transferability, uncertainty-aware field forecasting, adaptive spatial representations under changing source coverage, and latency-aware operational forecasting.

\section{Conclusion}
\label{sec:conclusion}

This study introduces an explicit regional PV generation field as a forecastable spatiotemporal representation of distributed PV generation. By organizing measurements from a distributed source fleet into a spatially structured state, the proposed SatPV-Field framework establishes a regional-to-local forecasting approach that uses predicted regional generation dynamics to inform individual-station predictions. The framework combines satellite-derived future shortwave radiation forecasts, which provide large-scale atmospheric information, with historical PV generation fields constructed from distributed station measurements and meteorological forecasts to predict future regional generation states. Experiments involving 496 distributed PV systems demonstrate the feasibility of this formulation, achieving a regional field forecasting MAE of 0.0880. The predicted future PV fields provide effective location-specific guidance for individual-station forecasting, reducing MAE by approximately 10.0\% relative to the best-performing single-station temporal baseline. A matched source-fleet comparison further shows that the field-based regional forecasting pathway achieves a 2.5\% lower MAE than its graph-based counterpart.

These findings demonstrate that distributed PV fleets can serve as regional sensing resources whose measurements support the explicit representation and prediction of evolving generation conditions. By connecting forecasted regional states with location-specific predictions, the proposed formulation establishes a regional-to-local forecasting pathway that extends beyond conventional station-centered modeling. More broadly, representing distributed generation as a spatial field provides a foundation for developing richer spatiotemporal models and exploring regional generation dynamics in distributed PV systems.

\section*{Acknowledgments}
This work was supported by the RISUD Project (P0042845), RISE Project (P0051003), and International Centre of Urban Energy Nexus (P0047700) of The Hong Kong Polytechnic University; the KKS Synergy project ``Energy flexibility through synergies of big data, novel technologies \& systems, and innovative markets'' (202000739); the Major Science and Technology Projects of Ningbo (2022Z236); and the Natural Science Foundation of Ningbo (2023J027). Computational support was provided by the High Performance Computing Centers at Eastern Institute of Technology, Ningbo, and Ningbo Institute of Digital Twin.



 
%

\bibliographystyle{IEEEtran}
\bibliography{sample}

@article{transformer,
  title={Attention is all you need},
  author={Vaswani, Ashish and Shazeer, Noam and Parmar, Niki and Uszkoreit, Jakob and Jones, Llion and Gomez, Aidan N and Kaiser, {\L}ukasz and Polosukhin, Illia},
  journal={Advances in neural information processing systems},
  volume={30},
  year={2017}
}

@article{multisite1,
  title={Improving multi-site photovoltaic forecasting with relevance amplification: DeepFEDformer-based approach},
  author={Wen, Yan and Pan, Su and Li, Xinxin and Li, Zibo and Wen, Wuzhenghong},
  journal={Energy},
  volume={299},
  pages={131479},
  year={2024},
  publisher={Elsevier}
}

@ARTICLE{multisite2,
  author={Simeunović, Jelena and Schubnel, Baptiste and Alet, Pierre-Jean and Carrillo, Rafael E.},
  journal={IEEE Transactions on Sustainable Energy}, 
  title={Spatio-Temporal Graph Neural Networks for Multi-Site PV Power Forecasting}, 
  year={2022},
  volume={13},
  number={2},
  pages={1210-1220},
  doi={10.1109/TSTE.2021.3125200}}

@article{new_ref3,
  title={Site-specific adjustment of a NWP-based photovoltaic production forecast},
  author={B{\"o}{\"o}k, Herman and Lindfors, Anders V},
  journal={Solar Energy},
  volume={211},
  pages={779--788},
  year={2020},
  publisher={Elsevier}
}

@article{chen2025selecting,
  title={Selecting effective NWP integration approaches for PV power forecasting with deep learning},
  author={Chen, Dayin and Shi, Xiaodan and Jiang, Mingkun and Zhu, Shibo and Zhang, Haoran and Zhang, Dongxiao and Chen, Yuntian and Yan, Jinyue},
  journal={Solar Energy},
  volume={301},
  pages={113939},
  year={2025},
  publisher={Elsevier}
}

@article{GEIT_chen,
title = {Learning to Forecast for Better Scheduling: A Regret-Aware Framework for PV-BESS Optimization},
journal = {Green Energy and Intelligent Transportation},
pages = {100385},
year = {2025},
issn = {2773-1537},
doi = {https://doi.org/10.1016/j.geits.2025.100385},
url = {https://www.sciencedirect.com/science/article/pii/S2773153725001355},
author = {Dayin Chen and Guotao Wang and Xiaodan Shi and Mingkun Jiang and Shibo Zhu and Haoran Zhang and Dongxiao Zhang and Yuntian Chen and Jinyue Yan}
}

@article{paper5,
  title={Ultra-short-term forecasting of large distributed solar PV fleets using sparse smart inverter data},
  author={Yue, Han and Ali, Musaab Mohammed and Lin, Yuzhang and Liu, Hongfu},
  journal={IEEE transactions on sustainable energy},
  volume={15},
  number={3},
  pages={1968--1980},
  year={2024},
  publisher={IEEE}
}

@inproceedings{multimodal_review4,
  title={Weather foundation model enhanced decentralized photovoltaic power forecasting through spatio-temporal knowledge distillation},
  author={He, Fang and Fan, Jiaqi and Deng, Yang and Zhang, Xiaoyang and Lau, Ka Tai and Wang, Dan},
  booktitle={Proceedings of the Thirty-Fourth International Joint Conference on Artificial Intelligence},
  pages={9665--9673},
  year={2025}
}

@inproceedings{PV_field_first,
  title={Solar PV maps for estimation and forecasting of distributed solar generation},
  author={de Hoog, Julian and Perera, Maneesha and Bandara, Kasun and Senanayake, Damith and Halgamuge, Saman},
  booktitle={ICML 2021 Workshop on Tackling Climate Change with Machine Learning},
  year={2021}
}

@techreport{R1,
  author      = {de l'Epine, Melodie and Dominguez, Guillermo and
                 Kaizuka, Izumi and J{\"a}ger-Waldau, Arnulf},
  title       = {A Snapshot of Global {PV} Markets},
  institution = {{IEA Photovoltaic Power Systems Programme, Task 1}},
  number      = {T1-49:2026},
  year        = {2026},
  month       = apr,
  doi         = {10.69766/WXLW1389},
  isbn        = {978-1-923734-04-3},
  url         = {https://iea-pvps.org/snapshot-reports/snapshot-2026/}
}

@misc{R2,
  author       = {{National Energy Administration of China}},
  title        = {Renewable Energy Grid-Connected Operation in 2025},
  year         = {2026},
  month        = feb,
  day          = {12},
  howpublished = {National Energy Administration of China},
  url          = {https://www.nea.gov.cn/20260212/742b8c6a078347b0b39de676c05c5d58/c.html},
  urldate      = {2026-07-11}
}

@article{erdener2022review,
  title={A review of behind-the-meter solar forecasting},
  author={Erdener, Burcin Cakir and Feng, Cong and Doubleday, Kate and Florita, Anthony and Hodge, Bri-Mathias},
  journal={Renewable and Sustainable Energy Reviews},
  volume={160},
  pages={112224},
  year={2022},
  publisher={Elsevier}
}

@article{R8,
  title={Comparing global and regional downscaled NWP models for irradiance and photovoltaic power forecasting: ECMWF versus AROME},
  author={Mayer, Martin J{\'a}nos and Yang, Dazhi and Szintai, Bal{\'a}zs},
  journal={Applied Energy},
  volume={352},
  pages={121958},
  year={2023},
  publisher={Elsevier}
}

@inproceedings{R7,
  title={A Preliminary Study on Regional PV Output Prediction Based on Spatial Downscaling},
  author={Li, Chengdong and Gong, Bingkuan and Cui, Chongyu and Zang, Kun and Quan, Xuezheng and Liu, Shuye},
  booktitle={2025 5th International Conference on Electrical Engineering and Control Science (IC2ECS)},
  pages={311--318},
  year={2025},
  organization={IEEE}
}

@article{R9,
  title={Spatio-temporal PV forecasting sensitivity to modules’ tilt and orientation},
  author={e Silva, R Amaro and Brito, MC},
  journal={Applied Energy},
  volume={255},
  pages={113807},
  year={2019},
  publisher={Elsevier}
}

@article{R9_2,
  title={A review of operational factors affecting photovoltaic system performance},
  author={Sep{\'u}lveda-Oviedo, Edgar Hernando},
  journal={Energy Conversion and Management: X},
  volume={26},
  pages={100942},
  year={2025},
  publisher={Elsevier}
}

@article{R11_1,
  title={Very short-term forecasting of distributed PV power using GSTANN},
  author={Yao, Tiechui and Wang, Jue and Wang, Yangang and Zhang, Pei and Cao, Haizhou and Chi, Xuebin and Shi, Min},
  journal={CSEE Journal of Power and Energy Systems},
  volume={10},
  number={4},
  pages={1491--1501},
  year={2022},
  publisher={CSEE}
}

@article{R11_2,
  title={Spatiotemporal federated learning based regional distributed PV ultra-short-term power forecasting method},
  author={Wang, Yuqing and Fu, Wenjie and Chen, Junfa and Wang, Junlong and Zhen, Zhao and Wang, Fei and Xu, Fei and Dui{\'c}, Neven and Yang, Di and Lv, Yuntong},
  journal={IEEE Transactions on Industry Applications},
  volume={60},
  number={5},
  pages={7413--7425},
  year={2024},
  publisher={IEEE}
}

@article{hersbach2020era5,
  title={The {ERA5} global reanalysis},
  author={Hersbach, Hans and Bell, Bill and Berrisford, Paul and Hirahara, Shoji and Hor{\'a}nyi, Andr{\'a}s and Mu{\~n}oz-Sabater, Joaqu{\'i}n and Nicolas, Julien and Peubey, Carole and Radu, Raluca and Schepers, Dinand and Simmons, Adrian and Soci, Cornel and Abdalla, Saleh and Abellan, Xavier and Balsamo, Gianpaolo and Bechtold, Peter and Biavati, Gionata and Bidlot, Jean and Bonavita, Massimo and De Chiara, Giovanna and Dahlgren, Per and Dee, Dick and Diamantakis, Michail and Dragani, Rossana and Flemming, Johannes and Forbes, Richard and Fuentes, Manuel and Geer, Alan and Haimberger, Leopold and Healy, Sean and Hogan, Robin J. and H{\'o}lm, El{\'i}as and Janiskov{\'a}, Marta and Keeley, Sarah and Laloyaux, Patrick and Lopez, Philippe and Radnoti, Gabor and de Rosnay, Patricia and Rozum, Iryna and Vamborg, Freja and Villaume, Sebastien and Th{\'e}paut, Jean-No{\"e}l},
  journal={Quarterly Journal of the Royal Meteorological Society},
  volume={146},
  number={730},
  pages={1999--2049},
  year={2020},
  doi={10.1002/qj.3803}
}

@misc{jaxa_ptree_himawari,
  author={{Japan Aerospace Exploration Agency}},
  title={{JAXA Himawari Monitor (P-Tree System): User's Guide}},
  year={2025},
  howpublished={JAXA Earth Observation Research Center},
  url={https://www.eorc.jaxa.jp/ptree/userguide.html},
  note={Accessed: 2026-07-17}
}

@article{bessho2016introduction,
  title={An introduction to Himawari-8/9—Japan’s new-generation geostationary meteorological satellites},
  author={Bessho, Kotaro and Date, Kenji and Hayashi, Masahiro and Ikeda, Akio and Imai, Takahito and Inoue, Hidekazu and Kumagai, Yukihiro and Miyakawa, Takuya and Murata, Hidehiko and Ohno, Tomoo and others},
  journal={Journal of the Meteorological Society of Japan. Ser. II},
  volume={94},
  number={2},
  pages={151--183},
  year={2016},
  publisher={Meteorological Society of Japan}
}

@article{new2,
  title={Large-Scale Distributed Photovoltaic Power Forecasting With Incomplete Data Based on Mixture of Experts Ensemble},
  author={Ding, Zhaohao and Duan, Xinyu and Zhao, Zhenkun and Wang, Yehan and Sun, Yimeng and Wei, Xuan},
  journal={IEEE Transactions on Industry Applications},
  year={2026},
  publisher={IEEE}
}

@article{new5,
  title={A novel approach for modeling spatial and temporal correlations of PV generation applied to prosumer virtual power plants},
  author={Power, Thomas and Verbi{\v{c}}, Gregor},
  journal={IEEE transactions on smart grid},
  volume={16},
  number={1},
  pages={183--193},
  year={2024},
  publisher={IEEE}
}

@article{new7,
  title={Ultra-short-term distributed PV power forecasting for virtual power plant considering data-scarce scenarios},
  author={Wang, Yuqing and Fu, Wenjie and Wang, Junlong and Zhen, Zhao and Wang, Fei},
  journal={Applied Energy},
  volume={373},
  pages={123890},
  year={2024},
  publisher={Elsevier}
}

@article{new3,
  title={Graph-based large scale probabilistic PV power forecasting insensitive to space-time missing data},
  author={Song, Keunju and Kim, Minsoo and Kim, Hongseok},
  journal={IEEE Transactions on Sustainable Energy},
  volume={16},
  number={1},
  pages={160--173},
  year={2024},
  publisher={IEEE}
}

@article{new10,
  title={Intra-hour irradiance forecasting techniques for solar power integration: A review},
  author={Chu, Yinghao and Li, Mengying and Coimbra, Carlos FM and Feng, Daquan and Wang, Huaizhi},
  journal={Iscience},
  volume={24},
  number={10},
  year={2021},
  publisher={Elsevier}
}

@article{new11,
  title={Impact of the spatio-temporal mismatch between satellite and in situ measurements on validations of surface solar radiation},
  author={Urraca, Ruben and Lanconelli, Christian and Gobron, Nadine},
  journal={Journal of Geophysical Research: Atmospheres},
  volume={129},
  number={10},
  pages={e2024JD041007},
  year={2024},
  publisher={Wiley Online Library}
}

@article{new8,
  title={Intraday spatiotemporal PV power prediction at national scale using satellite-based solar forecast models},
  author={Lanzilao, Luca and Meyer, Angela},
  journal={Energy and AI},
  pages={100786},
  year={2026},
  publisher={Elsevier}
}

@article{sup1,
  title={Intra-hour forecasts of solar power production using measurements from a network of irradiance sensors},
  author={Lonij, Vincent PA and Brooks, Adria E and Cronin, Alexander D and Leuthold, Michael and Koch, Kevin},
  journal={Solar energy},
  volume={97},
  pages={58--66},
  year={2013},
  publisher={Elsevier}
}

@article{sup2,
  title={Interpretable temporal-spatial graph attention network for multi-site PV power forecasting},
  author={Simeunovi{\'c}, Jelena and Schubnel, Baptiste and Alet, Pierre-Jean and Carrillo, Rafael E and Frossard, Pascal},
  journal={Applied Energy},
  volume={327},
  pages={120127},
  year={2022},
  publisher={Elsevier}
}

@article{sup3,
  title={Multi-site photovoltaic forecasting exploiting space-time convolutional neural network},
  author={Jeong, Jaeik and Kim, Hongseok},
  journal={Energies},
  volume={12},
  number={23},
  pages={4490},
  year={2019},
  publisher={MDPI}
}

@article{sup4,
  title={Forecasting the aggregated output of a large fleet of small behind-the-meter solar photovoltaic sites},
  author={Shaker, Hamid and Manfre, Daniel and Zareipour, Hamidreza},
  journal={Renewable Energy},
  volume={147},
  pages={1861--1869},
  year={2020},
  publisher={Elsevier}
}

\vfill

\end{document}